\documentclass[final]{cvpr}

\usepackage{times}
\usepackage{epsfig}
\usepackage{graphicx}
\usepackage{amsmath}
\usepackage{amssymb}
\usepackage{soul}

\usepackage[pagebackref=true,breaklinks=true,colorlinks,bookmarks=false]{hyperref}

\newcommand{\keypoint}[1]{\vspace{0.2cm}\noindent\textbf{#1}\quad}

\newcommand{\ayandas}[1]{#1}

\newcommand{\quickdraw}{\emph{Quick, Draw!}}
\newcommand{\cut}[1]{}

\DeclareMathOperator*{\argmin}{arg\,min}

\def\cvprPaperID{4018}
\def\confYear{CVPR 2021}

\begin{document}

\title{Cloud2Curve: Generation and Vectorization of Parametric Sketches}

\author{
    Ayan Das\textsuperscript{1,2}\quad\quad Yongxin Yang\textsuperscript{1,2}\quad\quad Timothy Hospedales\textsuperscript{1,3}\quad\quad Tao Xiang\textsuperscript{1,2}\quad\quad Yi-Zhe Song\textsuperscript{1,2}\\ \\
    \textsuperscript{1} SketchX, CVSSP, University of Surrey, United Kingdom\\
    \textsuperscript{2} iFlyTek-Surrey Joint Research Centre on Artificial Intelligence\\
    \textsuperscript{3} University of Edinburgh, United Kingdom\\
    {\tt\small \{a.das, yongxin.yang, t.xiang, y.song\}@surrey.ac.uk, t.hospedales@ed.ac.uk}
}

\maketitle

\begin{abstract}
Analysis of human sketches in deep learning has advanced immensely through the use of waypoint-sequences rather than raster-graphic representations. We further aim to model sketches as a sequence of low-dimensional parametric curves. To this end, we propose an inverse graphics framework capable of approximating a raster or waypoint based stroke encoded as a point-cloud with a variable-degree B\'ezier curve. Building on this module, we present Cloud2Curve, a generative model for scalable high-resolution vector sketches that can be trained end-to-end using point-cloud data alone. As a consequence, our model is also capable of deterministic vectorization which can map novel raster or waypoint based sketches to their corresponding  high-resolution scalable B\'ezier equivalent. We evaluate the generation and vectorization capabilities of our model on Quick, Draw! and K-MNIST datasets.
\end{abstract}

\section{Introduction}

The analysis of free-hand sketches using deep learning \cite{pengxu_survey} has flourished over the past few years, with sketches now being well analysed from classification \cite{sketchanet1,sketchanet2} and retrieval \cite{sketchx_fgsbir_1,sketchx_fgsbir_2,bhunia2020sketch} perspectives. 
Sketches for digital analysis have always been acquired in two primary modalities - \emph{raster} (pixel grids) and \emph{vector} (line segments). Raster sketches have mostly been the modality of choice for sketch recognition and retrieval \cite{sketchanet1,sketchx_fgsbir_1}. However, generative sketch models began to advance rapidly \cite{ha2017neural} after focusing on vector representations and generating sketches as sequences  \cite{bowman2016generating,srivastava2015unsupervised} of waypoints/line segments, similarly to how humans sketch. As a happy byproduct, this paradigm leads to clean and blur-free image generation as opposed to direct raster-graphic generations \cite{dcgan}.
Recent works have studied creativity in sketch generation \cite{ha2017neural}, learning to sketch raster photo input images  \cite{song2018learn2Sketch}, learning efficient human sketching strategies \cite{bhunia2020pixelor}, exploiting sketch generation for photo representation learning \cite{wang2020sketchembednet}, and the interaction between sketch generation and language \cite{huang2020scones}.



\begin{figure}
    \centering
    \includegraphics[width=\linewidth]{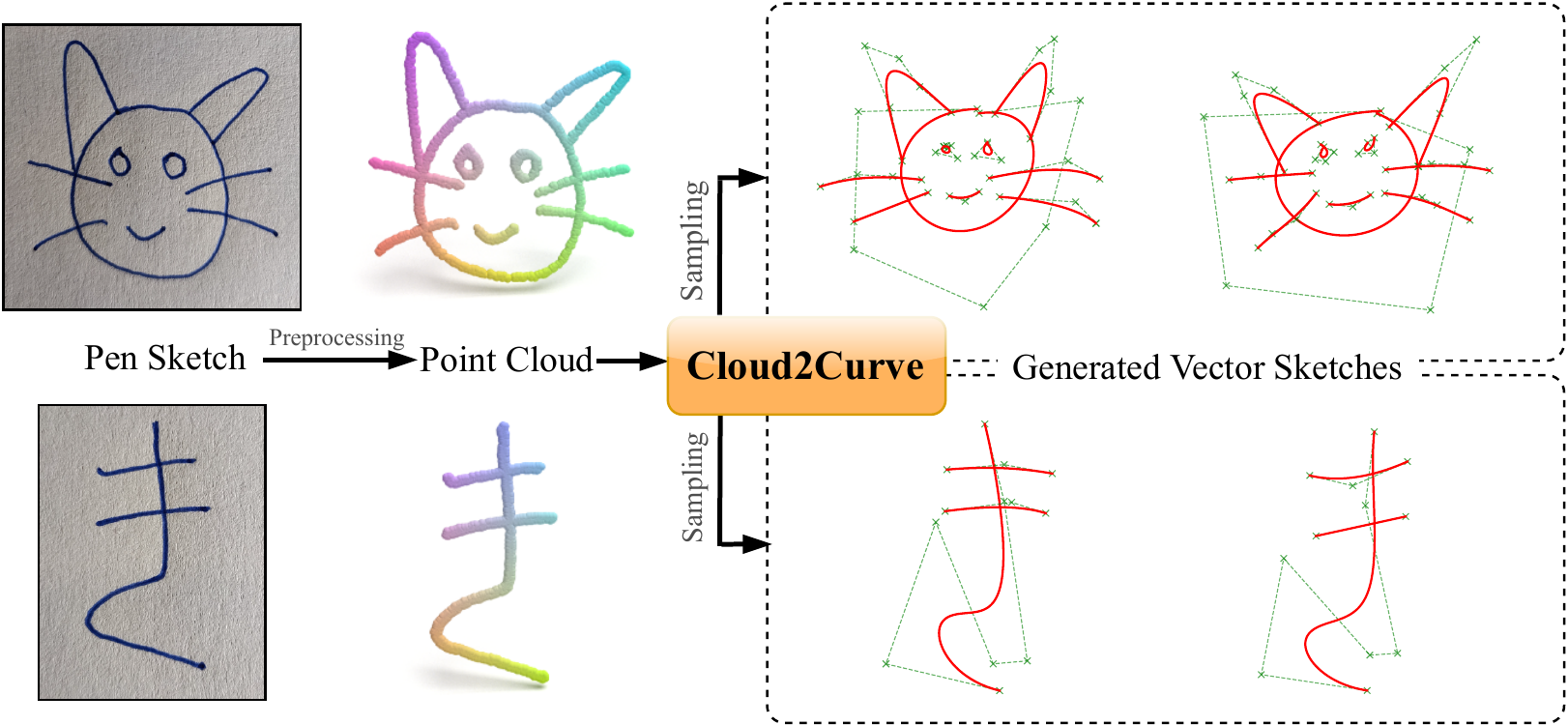}
    \caption{Cloud2Curve capability teaser. Top: Our trained model can vectorize a pen-on-paper sketch using scalable parametric curves. Bottom: Cloud2Curve can be trained on raster datasets such as KMNIST, where existing generative models cannot.}
    \label{fig:teaser}
\end{figure}

\ayandas{
We present Cloud2Curve, a framework that advances the generative sketch modeling paradigm on two major axes: (i) by generating \emph{parametric sketches}, i.e., a compositions of its constituent strokes as scalable parametric curves; and (ii) providing the ability to sample such sketches given point-cloud data, which is trivial to obtain either from raster-graphic or waypoint sequences. 
Altogether, our framework uniquely provides the ability to generate or deterministically vectorize scalable parametric sketches based on point clouds, as illustrated in Figure~\ref{fig:teaser} and explained below.
}

First, we note that although existing frameworks like SketchRNN \cite{ha2017neural} and derivatives generate vector sketches, they do so via generation of a dense sequence of short straight line segments. \ayandas{Consequently, (i) the output sketches are not \emph{spatially scalable} as required, e.g., for digital art applications, (ii) they struggle to generate \emph{long}  sketches due to the use of recurrent generators \cite{pascanu2013difficulty}, and (iii) the generative process suffers from low interpretability compared to human sketching, where the human mental model relies more on composing sketches with smooth strokes.} While there have been some initial attempts at scalable vector sketch generative models \cite{das2020bziersketch}, they were limited by the need for a two step training process, and to generate parametric B\'ezier curves of a fixed complexity, which poorly represent the diverse kinds of strokes in human sketches. In contrast, we introduce a machinery to dynamically determine the optimal complexity of each generated curve by introducing a continuous \emph{degree} parameter; and our model can further be trained end-to-end.

\ayandas{Second, existing generative sketch frameworks require purely sequential data in order to train their sequence-to-sequence encoder-decoder modules. This necessitates the usage of specially collected sequential datasets like \quickdraw{} and not general purpose raster sketch datasets. In contrast, we introduce a framework that encodes point-cloud training data and decodes a sequence of parametric curves. We achieve this through an inverse-graphics \cite{kulkarni2015vig,romaszko2017vig} approach of reconstructing training images/clouds by rendering its constituent strokes individually through a white-box B\'ezier decoder over several time-steps. To train this framework we compare reconstructed curves with the original segmented point cloud strokes with an Optimal Transport (OT) based objective. We show that such objectives, when coupled with appropriate regularizers, can lead to \emph{controllable abstraction} of sketches.
}

To summarize our contributions: \textbf{1)} We introduce a novel formulation of B\'ezier curves with a continuous degree parameter that is automatically inferred for each individual stroke of a sketch. \textbf{2)} \ayandas{We develop Cloud2Curve, a generative model capable of training and inference on point cloud data to produce spatially scalable \emph{parametric sketches}}. \textbf{(3)} We demonstrate scalable parametric curve generation from point-clouds, using \quickdraw{} and a subset of K-MNIST \cite{kmnistdataset} datasets.

\section{Related Works}

\keypoint{Generative Models} Generative models have been widely studied in machine learning literature \cite{bishop2006pattern} as a way of capturing complex data distributions. Probabilistic Graphical Models \cite{koller2009probabilistic} were hard to scale with variational methods \cite{blei2017variational}. With the advent of deep learning, neural approaches based on Variational Autoencoder (VAE) \cite{Kingma2013AutoEncodingVB} and Generative Adversarial Networks (GAN) \cite{GANgoodfellow14} were more scalable and able to generate high-quality images. The general formulation of VAE has been adapted to numerous problem settings. \cite{srivastava2015unsupervised} proposed the first model to encode sequential video data to a smooth latent space. Later, \cite{bowman2016generating} showed an effective way to train sequential VAEs for generating natural language. SketchRNN \cite{ha2017neural} followed a sequence model very similar to \cite{graves2013generating} and learned a smooth latent space for generating novel sketches. \cite{ganin2018imagePrograms} proposed a generative agent that learns to draw by exploring in the space of programs. Its sample inefficiency was ameliorated by \cite{zheng2018strokenet} through creating an environment model. More recently, the Transformer \cite{vaswani2017transformer} has been used to model sketches \cite{ribeiro2020sketchformer} due to the permutation invariant nature of strokes. 

\keypoint{Parametric representation} Although used heavily in computer graphics \cite{salomon2007curves}, parametric curves like B\'ezier, B-Splines, Hermite Splines have not been used much in mainstream Deep Learning. An early application of splines to model handwritten digits \cite{digit_splines_hinton} used a density model around b-splines and learns the parameters from a point-cloud using log-likelihood. B-Splines have been used as stroke-segments while representing handwritten characters with probabilistic programs \cite{lake_bpl}. SPIRAL \cite{ganin2018imagePrograms} is a generative agent that produces program primitives including cubic B\'ezier curve.
The font generation model in \cite{fontgen_iccv} and more recently DeepSVG Icon generator \cite{carlier2020deepsvg} treats fonts/icons as a sequence of SVG primitives. However, this requires the ground-truth SVG primitives. In contrast, we take an inverse graphics approach that learns to render point-clouds using parametric curves -- without any parametric curve ground-truth in the training pipeline. Stroke-wise embeddings are studied in \cite{aksan2020cose}, but this produces non-interpretable representations of each stroke, and still requires sequence data to train, unlike our inverse graphics approach. In summary, none of these methods can apply to raster data such as K-MNIST which we demonstrate here.

\keypoint{Learning parametric curves} The field of computer graphics \cite{salomon2007curves} has seen tremendous use of parametric curves \cite{deboor} in synthesizing graphics objects. However, parametric curves are still not popular in mainstream deep learning due to their usage of an extra latent parameter which is difficult to incorporate into a standard optimization setting. The majority of  algorithms for fitting B\'ezier \cite{cubicbezierfit,advancecubicbezierfit} or B-Splines \cite{splienfit_persample1,splinefit_persample2,bspline_lbfgs} are based on alternatively switching between optimizing control points and latent $t$ values which is computationally expensive, requires careful initialization and not suitable for pluging into larger computational graphs trained by backpropagation. Recently, B\'ezierEncoder \cite{das2020bziersketch} was proposed as a fitting method for B\'ezier curves by means of inference on any arbitrary deep recurrent model. We go beyond this to also infer curve degree, and fundamentally generalize it for training on point-cloud/raster data. 


\begin{figure}[t]
    \centering
    \includegraphics[width=\linewidth]{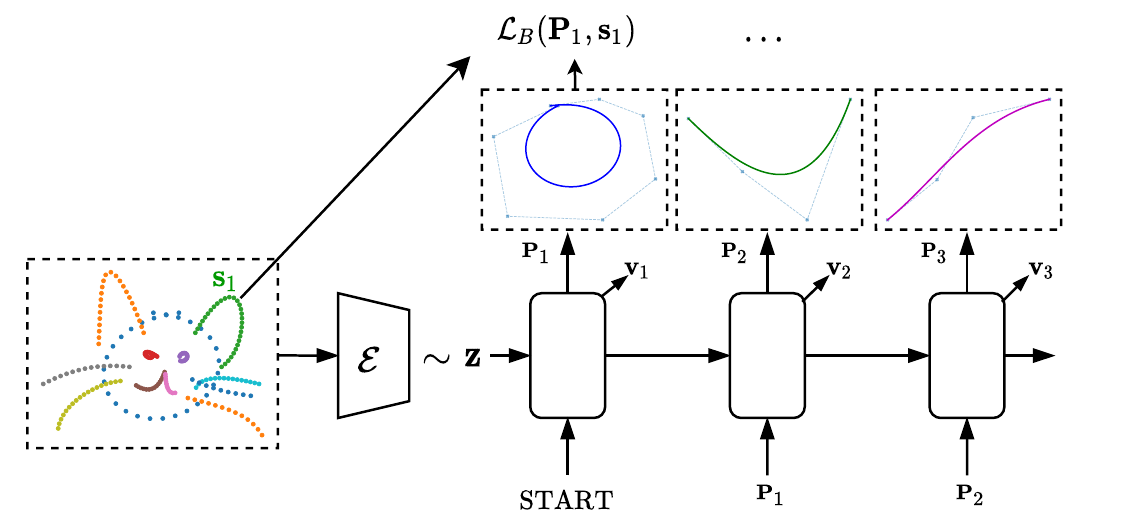}
    \caption{Overall diagram of the Cloud2Curve generative model.}
    \label{fig:overall}
\end{figure}

\section{Methodology}

The mainstream sketch representation popularized by \quickdraw{}, encodes a sketch as an ordered list of $L$ waypoints $[ (\mathbf{x}_i, \mathbf{q}_i) ]_{i=1}^L$ where $\mathbf{x}_i \in \mathbb{R}^2$ is the $i^{th}$ waypoint and $\mathbf{q}_i = ( q^{stroke}_i, q^{sketch}_i )$ is a tuple containing two binary variables denoting stroke and sketch termination. State-of-the-art sketch generation models such as SketchRNN \cite{ha2017neural} directly use this data structure for modelling the probability of a given sketch as the product of probabilities of individual waypoints given the previous waypoints

\vspace*{-0.3cm}
\begin{equation} \label{eq:sketchrnn}
    p_{sketchrnn}(\cdot) = \prod_{i=1}^L p(\mathbf{x}_i, \mathbf{q}_i \vert \mathbf{x}_{<i}, \mathbf{q}_{<i}; \theta)
\end{equation}

\noindent \ayandas{where the subscript $<i$ denotes the set of all indices before $i$. Our method treats sketches as a sequence of parametric curves, so we re-organize the data structure in terms of its strokes. We define a sketch $\mathcal{S} \triangleq [ \mathbf{S}_k ]_{k=1}^K$ as a sequence of its $K$ (may vary for each sketch) constituent strokes.
Some recent methods substitute the original $\mathbf{S}_k$ with separately learned non-interpretable transformer-based embedding \cite{aksan2020cose} or directly interpretable learned B\'ezier curves \cite{das2020bziersketch} with fixed degree. However, our method stands out due to the specific choice of variable degree B\'ezier curves as stroke embedding which is \emph{end-to-end trainable} along with the generative model.}

Furthermore, we simplify $\mathbf{S}_k$ by restructuring it into $(\mathbf{s}_k, \mathbf{v}_k)$ where $\mathbf{s}_k$ denotes a position-independent stroke and $\mathbf{v}_k$ is its position relative to an arbitrary but fixed point. This restructuring not only emphasizes the compositional relationship \cite{aksan2020cose} within the strokes but also helps the generative model learn position invariant parametric curves.

\subsection{Generative Model} \label{sec:genmodel}

Before defining our new variable-degree parametric stroke model in Section~\ref{sec:curverep} and the training objective in Section~\ref{sec:bezierloss}, we introduce the overall generative model.

\keypoint{Decoder/Generator} Unlike SketchRNN \cite{ha2017neural}, but similar to B\'ezierSketch \cite{das2020bziersketch} and CoSE \cite{aksan2020cose}, we model a sketch $\mathcal{S}$ autoregressively as a sequence of parametric strokes $\mathbf{P}_k$,
\begin{equation} \label{eq:ourmodel}
    p(\mathcal{S}) = \prod_{k=1}^K p(\mathbf{\mathbf{P}}_k, \mathbf{v}_k, \mathbf{q}_k \vert \mathbf{\mathbf{P}}_{<k}, \mathbf{v}_{<k}, \mathbf{q}_{<k} ; \theta)
\end{equation}
We use the binary random variable $\mathbf{q}_k$ to denote end-of-sketch as the usual way of terminating inference. Our model differs primarily in the fact that we model the density of the parametric stroke representation $\mathbf{P}_k$ at each step $k$. Since we do not use any pre-learned embedding as supervisory signal (unlike \cite{das2020bziersketch,aksan2020cose}), we do not have ground-truth for $\mathbf{P}_k$ and hence can not train it by directly maximizing the likelihood in Eq.~\ref{eq:ourmodel}. We instead minimize an approximate version of the negative log-likelihood:

\vspace*{-0.5cm}
\begin{equation} \label{eq:ourloss}
\begin{split}
    \mathcal{L}(\mathcal{S}) \approx &- \sum_{k=1}^K \left[ \mathcal{L}_B(\widehat{\mathbf{P}}_k, \mathbf{s}_k) + \log p(\mathbf{v}_k \vert \mathbf{\widehat{\mathbf{P}}}_{<k}, \mathbf{v}_{<k}) \right. \\
    &+ \left. \log p(\mathbf{q}_k \vert \mathbf{\widehat{\mathbf{P}}}_{<k}, \mathbf{v}_{<k}) \right], \\
    &\text{with } \widehat{\mathbf{P}}_k \sim p(\mathbf{P}_k \vert \mathbf{\widehat{\mathbf{P}}}_{<k}, \mathbf{v}_{<k})
\end{split}
\end{equation}

Instead of directly computing the log-likelihood of $\mathbf{P}_k$, we sample (re-parameterized) from the density and compute a downstream loss function $\mathcal{L}_B$ to act as a proxy. We  describe the exact form of $\mathbf{P}_k$ and $\mathcal{L}_B(\cdot)$ in Sections~\ref{sec:curverep}~\&~\ref{sec:bezierloss}.


\keypoint{Encoder} We condition the generation with a global latent vector \cite{ha2017neural}. This is produced by a VAE-style \cite{Kingma2013AutoEncodingVB} latent distribution whose parameters are computed using an encoder $\mathcal{E}_{\theta}$. A latent vector $\mathbf{z}$ is sampled as

\begin{equation} \label{eq:ourmodelwithkl}
    \mathbf{z}\vert\mathcal{S} \sim \mathcal{N}(\mathbf{\mu}, \mathbf{\Sigma})
    \text{ with } [\mathbf{\mu}, \mathbf{\Sigma}] = \mathcal{E}_{\theta}(\mathcal{S})
\end{equation}

\noindent by means of the reparameterization trick \cite{Kingma2013AutoEncodingVB}. $\mathbf{z}$ is then used to generate the mean and co-variance parameters at each step that define the distributions $p(\mathbf{P}_k\vert\cdot)$, $p(\mathbf{v}_k\vert\cdot)$ and $p(\mathbf{q}_k\vert\cdot)$ in Eq.~\ref{eq:ourloss}. A high level diagram of the full architecture is shown in Fig.~\ref{fig:overall}.

\keypoint{Training} Given our encoder and decoder, training is conducted by optimising the following objective

\begin{equation} \label{eq:ourlosswithkl}
    \sum_{\mathcal{S}\sim\mathcal{D}} \mathcal{L}(\mathcal{S}\vert\mathbf{z}) + w_{KL} \cdot \mathrm{KL}\left[ p_\theta ({\mathbf{z}\vert\mathcal{S}}) \vert\vert \mathcal{N}(\mathbf{0},\mathbf{\Sigma}) \right]
\end{equation}

Depending on the nature of $\mathcal{E}_{\theta}$, we can have very different kind of generative models. \ayandas{One can use the usual SketchRNN-style \cite{ha2017neural} encoder. But we use a Transformer based set encoder $\mathcal{E}_\theta(\mathcal{S})$ that can parse a given sketch as a cloud of ink-points and produce a concise latent vector representation.
In order to support learning parametric curve generation on point-cloud data, the remaining required components are a parametric curve model for $\mathbf{P}$ (Section~\ref{sec:curverep}) and  the loss $\mathcal{L}_B(\mathbf{P}_k,\mathbf{s}_k)$ describing how well a parametric stroke $\mathbf{P}_k$ fits the relevant point subset $\mathbf{s}_k$, described in Section~\ref{sec:bezierloss}}

\subsection{Representing variable degree B\'ezier curves} \label{sec:curverep}
The decoder generates the parameters of a B\'ezier curve representing a stroke at each step $k$. In this section we describe our new interpretable parametric curve representation $\mathbf{P}_k$. We formulate a flexible representation of $\mathbf{P}_k$ by defining it as a 2-tuple $(\mathcal{P}_k, r_k)$ comprised of: \textbf{1.} The parameters of a B\'ezier curve $\mathcal{P}_k \in \mathbb{R}^{(N+1)\times 2}$ where $N$ is the \emph{maximum allowable} degree, \textbf{2.} A continuous variable $r_k \in [0, 1]$ that will be used to determine the \emph{effective} degree of the B\'ezier curve. We  may drop the $k$ subscript when denoting an arbitrary time-step.

\keypoint{B\'ezier curves} \cite{salomon2007curves} are smooth and finite parametric curves used extensively in computer graphics. A degree $n$ B\'ezier curve is parameterized by $n+1$ \emph{control points}, and is usually modelled parametrically via an interpolation parameter $t\in[0,1]$. A B\'ezier curve with control points $\mathcal{P}$ can be instantiated using a pre-specified set of $G$ interpolation points $\displaystyle{ \left[ t_i \right]_{i=1}^G }$ as

\begin{equation} \label{eq:beziereqmatrix}
    C_{G\times 2} = \underbrace{ \begin{bmatrix}
        \cdots \\
        1, t_i, t_i^2, \cdots, t_i^n \\
        \cdots
    \end{bmatrix}}_{T_{G\times (n+1)}} \cdot M_{(n+1)\times (n+1)} \cdot \mathcal{P}_{(n+1) \times 2}
\end{equation}

\noindent where $G$ represents the granularity of rendering, which we treat as a hyperparameter. $T$ is the interpolation parameter matrix and $M$ is a matrix of Bernstein coefficients whose size and elements are dependent (only) on the degree $n$. For convenience, we will denote them as $M^{(n)}$. We next address how to use a continuous variable to induce an effective degree on $\mathcal{P}$.

\keypoint{Soft Binning} \cite{yang18dndt} has been introduced as a \emph{differentiable} way of binning a given real number into a predefined set of buckets. A real number $r$ needs to be tested against $n$ \emph{cut points} in order to assign a one-hot $(n+1)$-way categorical (or a continually relaxed) vector whose entries correspond to each of $n+1$ buckets/bins. Please refer to \cite{yang18dndt} for the detailed formulation of \emph{Soft Binning}.
We interpret each bucket as an \emph{effective degree} of a B\'ezier curve with its control points in $\mathcal{P}$. For our problem, we fix the cut points according to the maximum allowable degree $N$ as $\mathbf{U} = \displaystyle{ \bigl[ i/N \bigr]_{i=0}^N }$. This allows us to transform the continuous variable $r$ into a soft-categorical vector with $N+2$ components.
For practical benefit, we constraint the quantity $r$ to fall within the unit range of $[0, 1]$ by parameterizing it with an unconstrained variable $r' \in [-\infty, +\infty]$ as $r = \mathrm{Sigmoid}(r')$. With this added constraint, $r$ can only fall into $N$ buckets (avoiding the first and last open buckets), each of which may denote a B\'ezier curve with $2, 3, \cdots, N+1$ control points. Such design choice allows us to avoid representing a B\'ezier curve with 1 control point (which is invalid by definition) for any value of $r'$.

\begin{figure}
    \centering
    \includegraphics[width=0.9\linewidth]{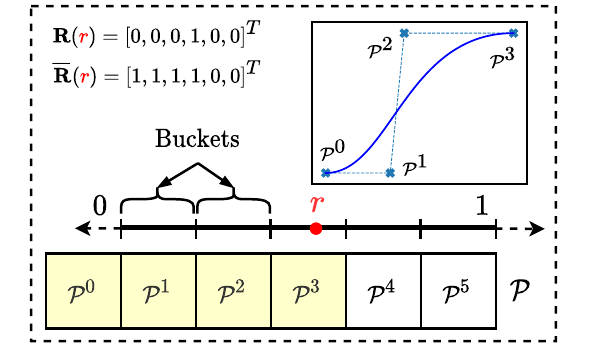}
    \caption{Visualization of the variable degree B\'ezier curve parameterized by its continuous degree parameter $r$.}
    \label{fig:vardegree}
\end{figure}

We define two quantities: \textbf{1)} A \emph{degree selector} $\mathbf{R}(r)$ whose element $\mathbf{R}_i$ is $1$ iff $r$ falls into the bin designated for degree $i$; \textbf{2)} A \emph{control points selector} $\overline{\mathbf{R}}(r)$ defined as the \emph{reversed cumulative summation} of $\mathbf{R}(r)$. Please refer to Fig.~\ref{fig:vardegree} for an illustration of a  variable degree B\'ezier curve.
\cut{For a concrete example, if the value of $r$ falls into $2^{nd}$ bin and $N=4$, then $\mathbf{R} = [0, 0, 1, 0, 0]^T$ and $\overline{\mathbf{R}} = [1, 1, 1, 0, 0]^T$. In the next section, we show how to use $\mathbf{R}$ and $\overline{\mathbf{R}}$.}



\keypoint{Variable degree} We can now augment $\mathcal{P}$ with $r$ to create variable degree B\'ezier representation. We mask interpolation parameters $T$ and control points $\mathcal{P}$ using the \emph{control point selector} as

\vspace*{-0.5cm}
\begin{equation*}
    \widehat{T}(r) = \underbrace{ \begin{bmatrix}
        \cdots \\
        1, t_i, t_i^2, \cdots, t_i^N \\
        \cdots
    \end{bmatrix}}_{T_{G\times (N+1)}}
    \odot
    \begin{bmatrix}
        \overline{\mathbf{R}}^T(r) \\
        : \\
        \overline{\mathbf{R}}^T(r)
    \end{bmatrix}_{G\times (N+1)}
\end{equation*}

\begin{equation*}
    \widehat{\mathcal{P}}(r) = \mathcal{P} \odot \begin{bmatrix}
        \overline{\mathbf{R}}(r), \overline{\mathbf{R}}(r)
    \end{bmatrix}
\end{equation*}

For $M$, we need to select the correct one from the set $\{ M^{(n)} \}_{n=0}^N$ according to the value of $r$. We accomplish this by first defining a $3D$ tensor $\mathcal{M} \in \mathbb{R}^{(N+1) \times (N+1)\times (N+1)}$ where

\begin{equation*}
\mathcal{M}[i,\cdots] = \begin{bmatrix}
    M^{(i)} & \mathbf{0} \\
    \mathbf{0} & \mathbf{0}
    \end{bmatrix}_{(N+1) \times (N+1)}
\end{equation*}

\noindent The $\mathbf{0}$s denote appropriately sized zero matrices used as fillers. We can then compute $\widehat{M}$ using the \emph{degree selector} as

\begin{equation*}
    \widehat{M}(r) = \mathbf{R}^T(r) \cdot \mathcal{M}
\end{equation*}

With all three components augmented with masks, we can write the variable degree version of Eq.~\ref{eq:beziereqmatrix} as a function of the degree parameter $r$ as

\begin{equation}
    \widehat{C}(r)_{G\times 2} = \widehat{T}(r) \cdot \widehat{M}(r) \cdot \widehat{\mathcal{P}}(r) \label{eq:finalvarrender}
\end{equation}


\subsection{Learning and Inference}\label{sec:bezierloss}

Given the generative model described in Section~\ref{sec:genmodel} and our new curve representation in Section~\ref{sec:curverep}, we now describe how to train on point-cloud data, and how to use our trained model to vectorize new point cloud inputs. 

\keypoint{B\'ezier Loss: Cloud} 
\cut{Given a curve $\mathbf{P}$ and stroke cloud $\mathbf{s}$, we n} 
The final component required by our generative model (Eq.~\ref{eq:ourloss}) is a loss  $\mathcal{L}_B(\mathbf{P}, \mathbf{s})$  to measure the similarity between a point cloud based stroke $\mathbf{s}$ and the curve  $\mathbf{P}$ . We discard the sequential information in the set $\widehat{C}$ and can compute any Optimal Transport (OT) based loss like EMD (Earth mover's distance) or Wasserstein Distance \cite{arjovsky17wgan}. The specific distance we used in our experiments is the Sliced Wasserstein Distance (SWD) \cite{kolouri2018sliced}:

\begin{equation} \label{eq:lossfunc_swd}
    \mathcal{L}_B(\mathbf{P}, \mathbf{s}) = \mathrm{SWD}(\mathbf{P}, \mathbf{s})
\end{equation}

Since OT-based losses are theoretically designed to measure the difference between two distributions, it is necessary to ensure the cardinality of the sets (either $\mathbf{P}$ or $\mathbf{s}$) are sufficiently high. We use a large enough $G$ for instantiating the B\'ezier curve and also densely resample $\mathbf{s}$ to the same granularity.

\keypoint{B\'ezier Loss: Sequence} 
If \emph{optional} sequence information for $\mathbf{s}$ is available, we can compute a point-to-point MSE loss:

\begin{equation*}
    \mathcal{L}_B(\mathbf{P}, \mathbf{s}) = \sum_{g=1}^G \vert\vert \widehat{C}^g(r) - \mathbf{s}^g \vert\vert_2.    
\end{equation*}

\noindent between each point $s^g$ on the curve and each interpolation point $\widehat{C}^g$ on the rendered curve $\mathbf{P}$ (Eq.~\ref{eq:finalvarrender}).

\begin{figure*}
    \centering
    \includegraphics[width=\linewidth]{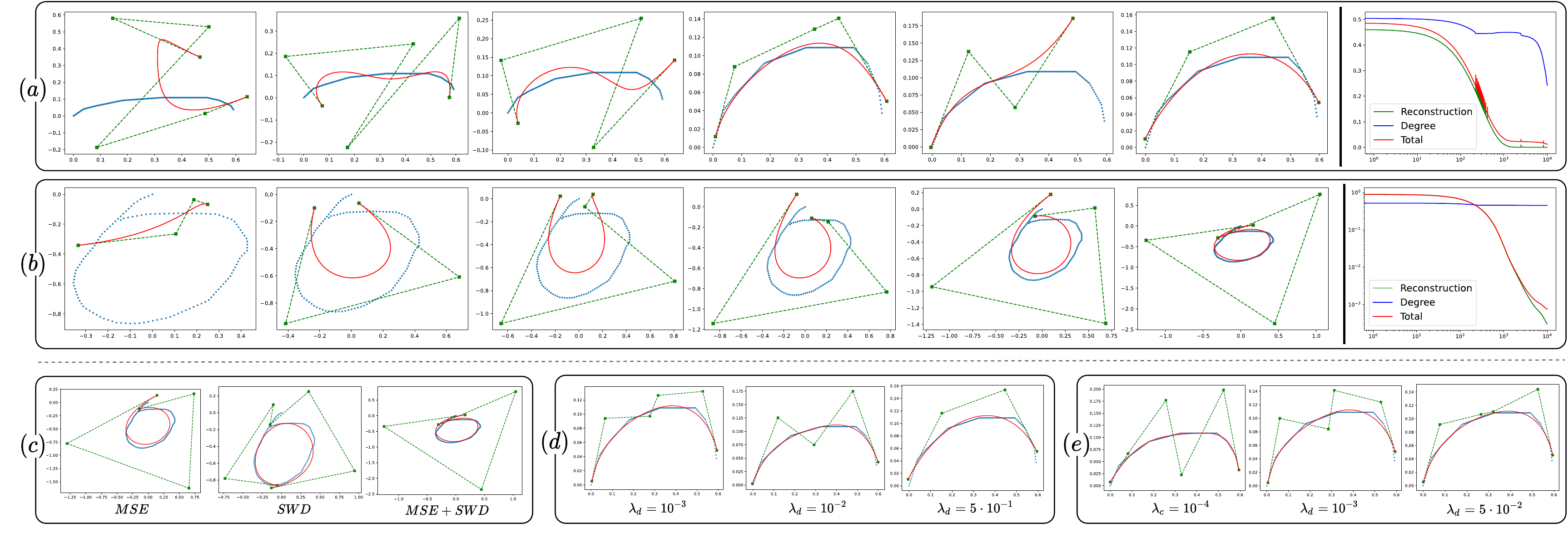}
    \caption{Visualization of fitting a variable degree B\'ezier curve (red) to individual strokes represented as point-clouds (blue). (a \& b) Two examples with corresponding training loss components on the right. The training iteration increases from left to right. (c) Different choices of loss (section \ref{sec:bezierloss}). (d) Different choices of degree regularizer $\lambda_d$. (e) Different choices of control point regularizer $\lambda_c$.}
    \label{fig:varbezqual}
\end{figure*}


\keypoint{Regularization} The purpose of introducing a variable degree B\'ezier curve formulation is to provide the model with flexibility to encode strokes with perfect fit. To avoid overfitting strokes by using a complex curve to fit a simple stroke, the learning phase should be provided with incentive to reduce the degree whenever possible. A simple regularizer on our degree variable $r$ could achieve this:

\begin{equation} \label{eq:loss_with_degreereg}
    \widehat{\mathcal{L}}_B(\mathbf{P}, \mathbf{s}) = \mathcal{L}_B(\mathbf{P}, \mathbf{s}) + \lambda_{d} \cdot r
\end{equation}

Apart from this, we also added another regularizer to reduce the level of overfitting given a degree. Overfitting in learning B\'ezier curve can occur when control points can move anywhere during the optimization. Following \cite{das2020bziersketch}, we add another term to the loss function to penalize the consecutive control points moving away from each other

\begin{equation} \label{eq:loss_with_both_reg}
\begin{split}
    \widehat{\mathcal{L}}_B(\mathbf{P}, \mathbf{s}) &= \mathcal{L}_B(\mathbf{P}, \mathbf{s}) + \lambda_d \cdot r \\
    &+ \lambda_{c} \cdot \left( \sum_{i=0}^{N} \vert\vert \mathcal{P}^{i+1} - \mathcal{P}^{i} \vert\vert_2 \right) \odot \overline{\mathbf{R}}(r)
\end{split}
\end{equation}

\noindent where we have masked out control points in $\mathcal{P}$ that are not meaningful given the degree value $r$.

\keypoint{Implementation Details} Our training objective  (Eq~\ref{eq:ourlosswithkl}) encodes whole images but fits one parametric curve at a time to the set of points corresponding to a stroke (Eq \ref{eq:loss_with_both_reg}). In principle all the strokes could be emitted by the generator, and then compute the loss between the full parametric sketch and the full point cloud. However for stability of optimisation, and limiting the cost of OT computation between curves and cloud, we proceed stroke-wise. To do this for genuine raster data, we pre-process the input point-cloud with 2D clustering to segment into strokes, and then iterate over the strokes in random order to train the model. Note that this priveleged information is only required during training, after training we can vectorize an unsegmented raster image into parametric curves, as shown in Figure~\ref{fig:teaser}.

\keypoint{Inference: Generation \& Vectorization} \ayandas{Given our trained model, we can use it for conditional generation. Given a sketch $\mathcal{S}$ as pointcloud, we simply sample a latent vector $\widehat{\mathbf{z}} \sim \mathcal{N}(\mathcal{E}_{\theta^*}(\mathcal{S}))$ following Eq.~\ref{eq:ourmodelwithkl} and use it to construct the parameters of the distributions $p(\mathbf{P}_k\vert\cdot)$, $p(\mathbf{v}_k\vert\cdot)$ and $p(\mathbf{q}_k\vert\cdot)$. We further sample}

\begin{equation*}
\begin{split}
    \widehat{\mathbf{P}}_k, \widehat{\mathbf{v}}_k, \widehat{\mathbf{q}}_k \sim p(\mathbf{P}_k\vert\cdot) \cdot p(\mathbf{v}_k\vert\cdot) \cdot p(\mathbf{q}_k\vert\cdot)
\end{split}
\end{equation*}

\noindent \ayandas{iteratively at each time-step and stop only when $\widehat{\mathbf{q}}_k$ is in \emph{end-of-sequence} state. To visualize, we simply render all the $(\widehat{\mathbf{P}}_k, \widehat{\mathbf{v}}_k)$ pair on a canvas.}


\ayandas{We can \emph{vectorize} a given sketch $\mathcal{S}$ deterministically by following a similar procedure as generation but with discarding the source of stochasticity while sampling. We can simply assign all the co-variance parameters of $p(\mathbf{P}_k\vert\cdot)$, $p(\mathbf{v}_k\vert\cdot)$ and $p(\mathbf{q}_k\vert\cdot)$ to zero.}


\section{Experiments}

\keypoint{Datasets} \quickdraw{} \cite{ha2017neural} is the largest free-hand sketch dataset available till date. \quickdraw{} is created by collecting drawings from a fixed set of categories drawn under a game played by millions all over the world. Although \quickdraw{} is collected on a vast array of digital devices like smartphone, tablets etc., the data acquisition technique is kept uniform. The sketches are collected as a series of 2D waypoints along the trajectory of ink flow. To demonstrate our model, we use \quickdraw{} but discard it's way-point sequence information, using the waypoints as point-cloud data. We use \quickdraw{} stroke-level segmentation (given by pen-up and pen-down indicators) as priveleged information during training. To demonstrate our model's ability to train on pure raster data, for which neither point-sequence nor stroke-sequence information is available, we also validate our model on a few classes ($0,1,8,9$) of K-MNIST \cite{kmnistdataset}. We extract a  point-cloud representation from K-MNIST by simple binarization and thinning. For training we segment into strokes by Spectral Clustering.


\subsection{Variable-degree B\'ezier curve}

We first validate the ability of our new curve representation to fit variable-degree B\'ezier curves  to isolated strokes represented as point clouds. 

\keypoint{Setup}  We collected few strokes from the sketches of \quickdraw{} \cite{ha2017neural} dataset. Since each stroke may have different number of $2D$ waypoints, we densely resample them uniformly with a fixed granularity of $G_0$. Hence, given a waypoint-based stroke $\mathbf{s} \in \mathbb{R}^{G_0\times 2}$, we directly optimize the B\'ezier curve parameters by means of Eq.~\ref{eq:loss_with_both_reg}
\begin{equation}
\begin{split}
    \mathcal{P}^{*}, r'^{*} &= \argmin_{\mathcal{P}, r'} \widehat{\mathcal{L}}_B(\mathcal{P}, r, \mathbf{s}),\text{ with}\\
    r &= \mathrm{Sigmoid}(r')\text{, and } (\mathcal{P}, r) \triangleq \mathbf{P}
\end{split}
\end{equation}

The gradients of the loss w.r.t parameters, i.e. $\displaystyle{ \left(\frac{\partial \mathcal{L}_B}{\partial \mathcal{P}}, \frac{\partial \mathcal{L}_B}{\partial r'} \right) }$ are computed simply by backpropagation and updated using any SGD-based algorithm.

Additionally, we experimented with point-to-point MSE loss as described in Section~\ref{sec:bezierloss} in cases where sequential information is available. Usage of privileged information helps learning the B\'ezier parameters quickly. However, the usage of \emph{only} MSE loss degrades the fitting quality. We noticed that the dense uniform resampling of strokes $\mathbf{s}$ do not have a proper point-to-point alignment with an instantiated B\'ezier curve with uniform set of $t$-values of same granularity $G$. An instantiated B\'ezier curve tends to have dense distribution of points in places of bending and sparse otherwise, while the point cloud data derived from a waypoint sequence does not. Using the Sliced Wasserstein Distance (SWD) \cite{kolouri2018sliced} with MSE loss proved most effective with the sequential information in the MSE loss helping speed of convergence, and the SWD loss helping quality of fit. 

\keypoint{Results} We show qualitative results of learning variable degree B\'ezier curves from point-cloud based strokes $\mathbf{s}$. Figure~\ref{fig:varbezqual}(a, b) shows two examples of learning the B\'ezier control points $\mathcal{P}$ and  degree parameter $r'$.
To interpret the learned value of $r$, we run  binning  on it using the pre-defined cut-points $\mathbf{U}$  in Section~\ref{sec:curverep} and retrieve the degree $n$.
In Fig.~\ref{fig:varbezqual}(a), we see the degree reduces from $n=5$ to $n=4$ and then $n=3$, since the curve is simple. Similarly, Fig.~\ref{fig:varbezqual}(b) shows an increase in degree due to the higher complexity stroke. The last figure in each example shows each loss component over iterations. We used $\lambda_d = 10^{-3}$, $\lambda_c = 5\cdot 10^{-2}$, $G_0=128$ and maximum degree $N=6$ for this experiment.
Fig.~\ref{fig:varbezqual}(c) shows the qualitative difference between learning with SWD, MSE alone and both combined. We see that SWD tends to ignore small details but preserves the overall structure.
Fig.~\ref{fig:varbezqual}(d, e) also shows the effect of both regularizer strengths  ($\lambda_c\text{ and }\lambda_d$). \ayandas{It is evident from the figure that $\lambda_d$ is responsible for pressuring the reduction of degree in fitting, whereas $\lambda_c$ reduces overfitting by keeping the control points close to each other.}

\begin{figure}
    {\centering
    \includegraphics[width=\linewidth]{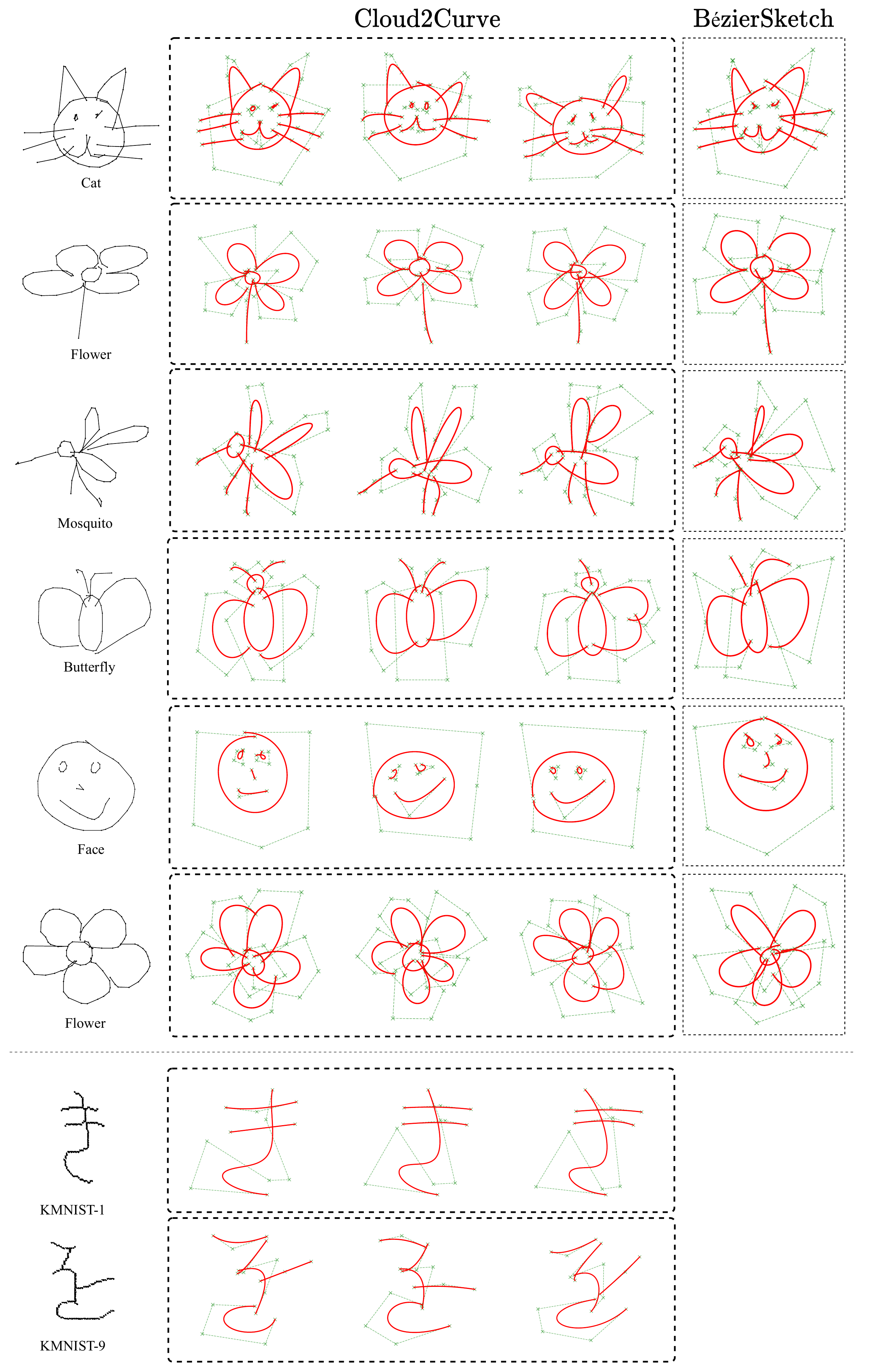}}
    \caption{\ayandas{Qualitative results for conditional generation on \quickdraw{} and a limited subset of KMNIST. Three vector sketches are generated by means of sampling from Cloud2Curve. For a qualitative comparison, we also provide one sample from B\'ezierSketch \cite{das2020bziersketch} for the same input sketch. Due to unavailability of purely sequential information, B\'ezierSketch cannot be trained on KMNIST.}}
    \label{fig:qual}
\end{figure}

\begin{figure}
    \centering
    \includegraphics[width=\linewidth]{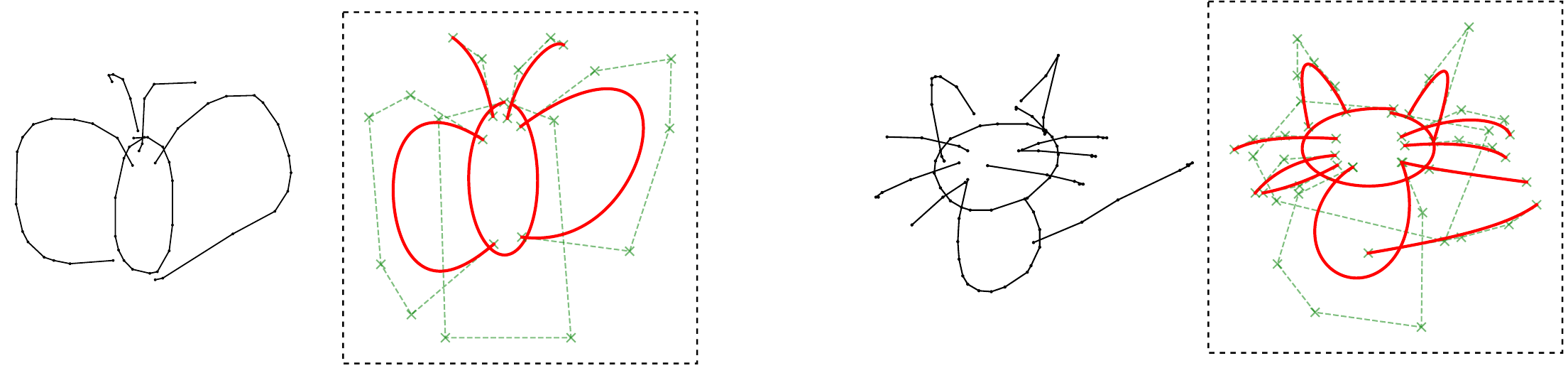}
    \caption{Deterministically vectorized sketches from the model trained with $\lambda_d = 10^{-3}$.}
    \label{fig:qual_vec}

\end{figure}

\subsection{Generation and \& Vectorization Model}

\keypoint{Setup} We pre-process the sketches in \quickdraw{}  and K-MNIST into a sequence of position-independent strokes and their starting points $\displaystyle{ \mathcal{S} = \bigl[ (\mathbf{s}_k, \mathbf{v}_k) \bigr]_{k=1}^K }$. To stabilize the training dynamics, we also center the whole sketch and scale it down numerically to fit inside a unit circle.  As part of data augmentation, we performed the following: (i)  We made sure the strokes do not have too sharp bends. We split a stroke into multiple strokes from a point with high curvature or when it is too long. \ayandas{Such transformation alleviates the problem of learning overly smooth representations up to a great extent.} (ii)  We added standard Gaussian noise to each $2D$ point in the stroke.

\keypoint{Implementation Details} 
Although our decoder model in Eq.~\ref{eq:ourmodel} is very generic, we chose to use a standard RNN. Alternatives such as Transformer \cite{ribeiro2020sketchformer} could also be used.

The consequence of using position independent strokes is that the predicted parametric curve must also be position independent.
\cut{We may leave it to the optimizer to learn that the first control point $\mathcal{P}^0$ is at origin.} 
A trick to guarantee that $\mathcal{P}^0 = [0, 0]^T$ is to only predict $N$ control points for $i=1\rightarrow N$ and explicitly fix $\mathcal{P}^0 := [0, 0]^T$.

We set the family of distributions $p(\mathbf{P}_k\vert \cdot; \theta)$ and $p(\mathbf{v}_k\vert \cdot)$ at each time-step as isotropic Gaussian with their mean and std vector predicted by our encoder $\mathcal{E}$. 
The reason we chose to not use GMM here is because of its difficulty to reparameterize (as required in Eq.~\ref{eq:ourloss}). However, the loss of expressiveness is compensated by increasing model capacity. 
Initiation of inference requires a special input state. Since $(\mathcal{P}, r, \mathbf{v}) = (\mathbf{0}, 0, \mathbf{0})$ is a semantically invalid state, it can be used as a special token to kickstart the generation and also use in teacher-forcing while training.

\cut{There are a number of choices for the encoder $\mathcal{E}_{\theta}$. A natural way of \emph{parsing} a given sketch is to encode the rasterized image using a  CNN  encoder. But since raster data is high dimensional and complex, a better alternative is to use Transformer encoders \cite{ribeiro2020sketchformer} on the point cloud representation of the sketch, as a better bridge between raster and vector representations.}
\ayandas{ As mentioned earlier, we use a set transformer as the encoder $\mathcal{E}_{\theta}$.
Specifically, we use the strategy of  \cite{ribeiro2020sketchformer} to compute a compact latent distribution by simply applying a learnable self-attention layer on the encoder features.
Note that, unlike the decoding side, we never use stroke level segmentation for the encoder. Thus we can perform inference without stroke-segmented sketches.}


For the decoder, we used a $2$-layer LSTM with hidden vectors of size $1024D$. We fixed the value of maximum allowable degree to $N=9$, therefore the B\'ezier curves can have at max $10$ control points which is more than enough to represent fairly complex geometry. For computing the B\'ezier loss $\mathcal{L}_B$, we used a granularity of $G=128$. For the transformer-based encoder model, we used a $512D$ transformer with $8$ heads and $4$ layers. For stable training, we follow the usual KL annealing trick \cite{bowman2016generating,ha2017neural} by varying $w_{KL}$ from $0$ to $1$ over epochs.

\begin{figure}[b]
    \centering
    \includegraphics[width=\linewidth]{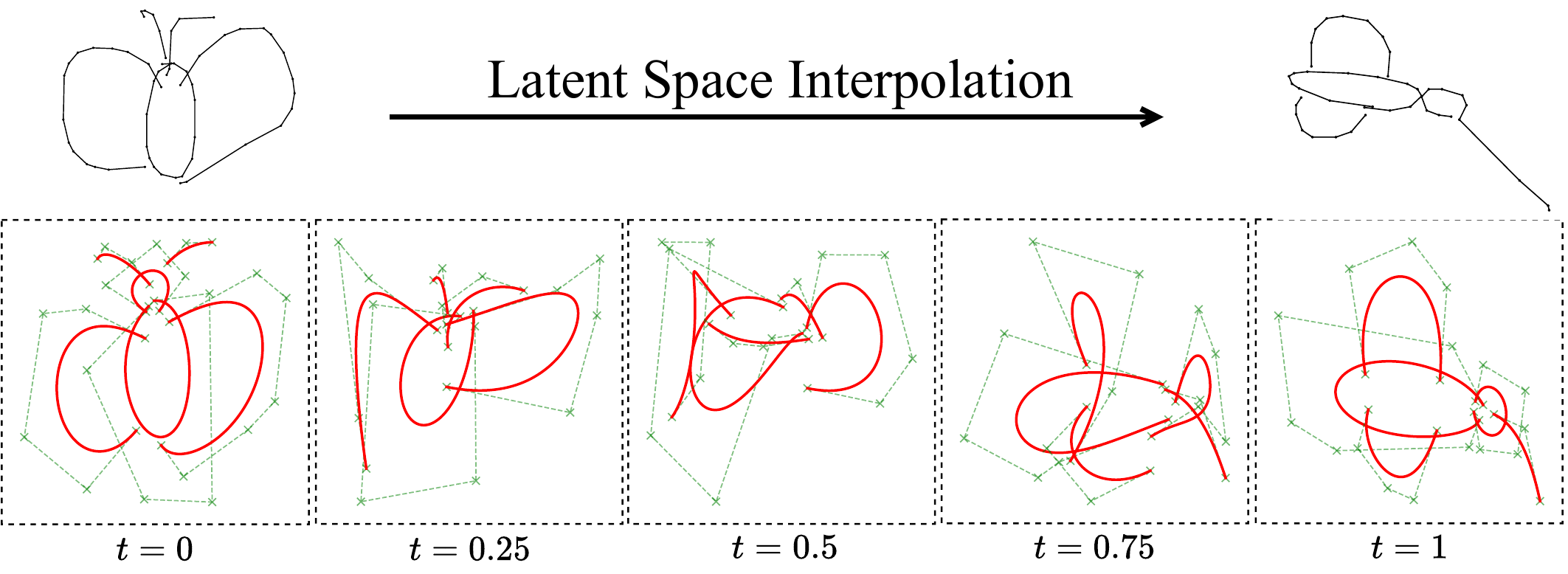}
    \caption{Interpolating between two sketches by walking on the latent space as: $\mathbf{z} = \mathbf{z}_{butterfly} \cdot (1-t) + \mathbf{z}_{mosquito} \cdot t$}
    \label{fig:latent_walk}
\end{figure}

\keypoint{Qualitative Results of Sketch Generation and Vectorization} 
We trained one model for each class from both \quickdraw{} and K-MNIST datasets. Since K-MNIST dataset is fairly small, we do not train a model from scratch. Instead, we finetune on a model pre-trained on \quickdraw{}. \ayandas{Fig.~\ref{fig:qual} shows qualitative results in terms of conditional generation from our generative sketch model. We also show qualitative results for deterministic raster-to-curve vectorizations in Fig.~\ref{fig:qual_vec}. Both results are computed following the procedures described in \ref{sec:bezierloss}. Trained with two categories, the model can also interpolate between two given sketches \cite{ha2017neural} by interpolating on the latent space (refer to Fig.~\ref{fig:latent_walk} for an example).}

\begin{figure}[t]
    \centering
    \includegraphics[width=\linewidth]{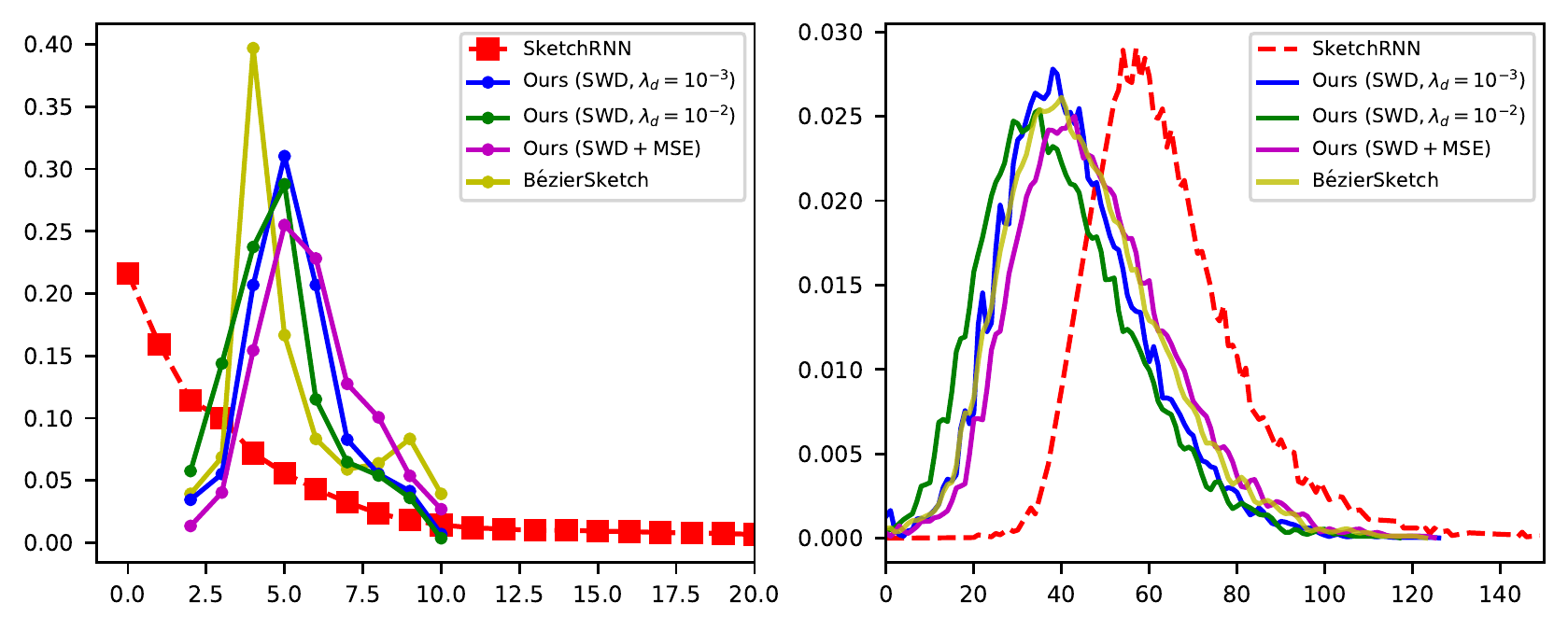}
    \includegraphics[width=\linewidth]{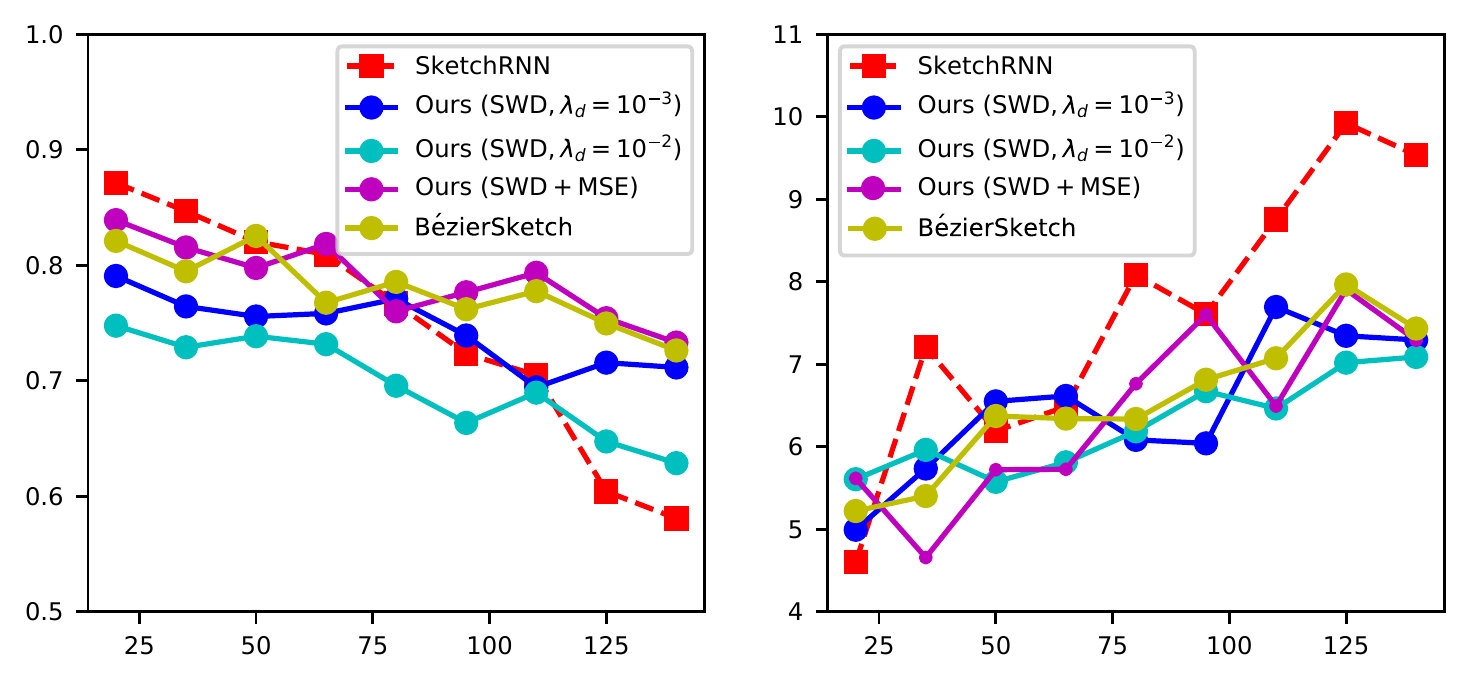}
    \caption{\ayandas{Comparison between Cloud2Curve and other generative models in terms of (Top left, a) stroke-histogram, (Top right, b) sketch-histogram, (Bottom left, c) Classification accuracy-vs-length and (Bottom right, d) FID-vs-length for the generated samples. Cloud2Curve generates more compact sketches and scales better to higher lengths in terms of FID and recognition accuracy.}}
    \label{fig:lenhist}
\end{figure}

\begin{figure}
    \centering
    \includegraphics[width=\linewidth]{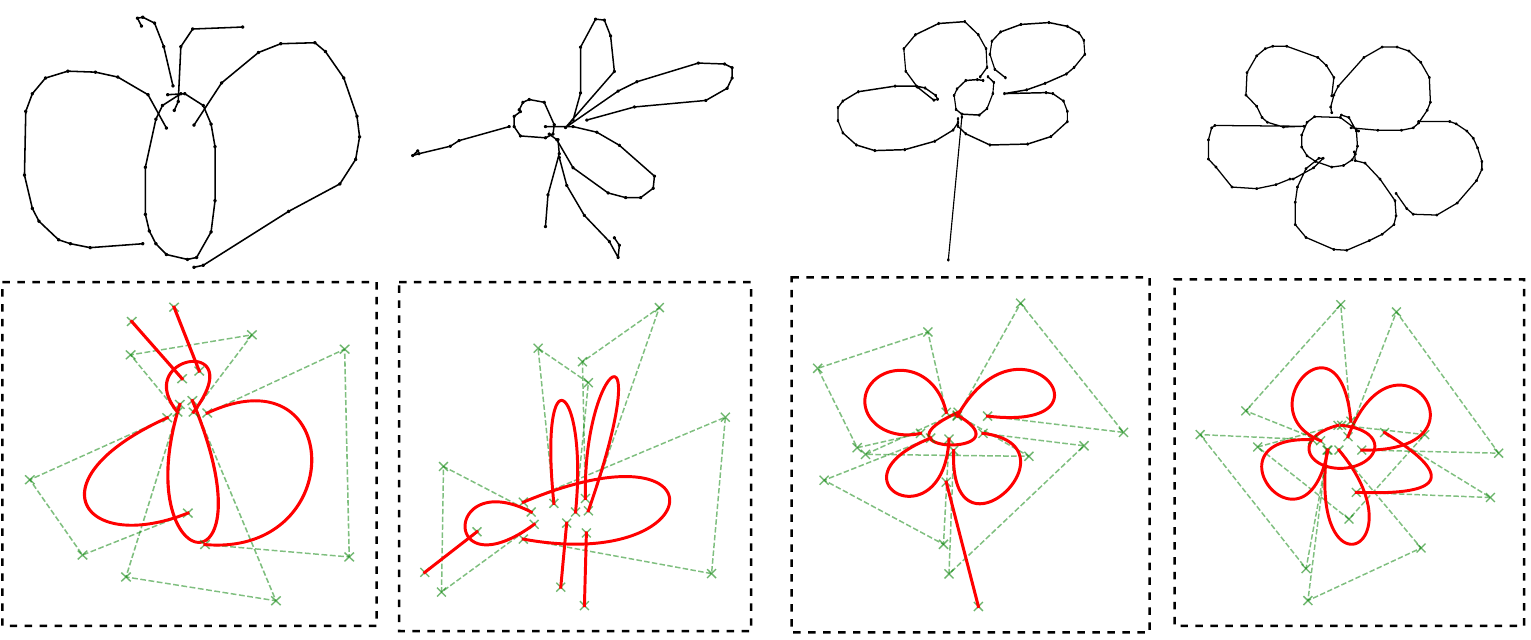}
    \caption{Input sketches and conditionally generated samples from a model trained with $\lambda_d = 10^{-2}$.}
    \vspace{-0.2cm}
    \label{fig:large_lambda_d}
\end{figure}

\keypoint{Cloud2Curve Generates Compact Sketches} 
One important technical benefit of generating sketches using B\'ezier curves is that the generated sketches are shorter in terms of number of points to be stored. Moreover, short representation of sketches lead to less complex downstream models like classification and retrieval. The length histograms both at stroke-level (points per stroke) and sketch-level (points-per-sketch) of generated sketches are shown in Figure.~\ref{fig:lenhist}(a,b) demonstrate this point for Cloud2Curve vs SketchRNN and B\'ezierSketch. We notice that the number of points per stroke is heavily concentrated around $5$ (and bounded by $2 \rightarrow 10$) for all our models but SketchRNN has a much larger spread. As a direct consequence, the average number of points in an entire sketch is also smaller (by a margin of $~20$). We also show that increasing $\lambda_d$ reduces the lengths even further, thereby achieving more \emph{abstract} sketches (Refer to Fig.~\ref{fig:large_lambda_d} for qualitative examples). Using MSE loss has an effect of increasing the length because it tries to fit the data more precisely considering small artifact.

\keypoint{Quantitative Analysis on Sketch Generation Quality}
To confirm that our generative model samples are semantically plausible, we rendered them and classified them using a CNN classifier \cite{sketchanet2} trained on real sketches from \quickdraw{}. The results in Figure~\ref{fig:lenhist}(c) show that our parametric curve sketches are indeed accurately recognizable by state-of-the-art image classifiers even when longer in length.

To further assess generation quality, the FID score \cite{fidscore} is also computed. We compute a modified FID score using rasterized samples after projecting them down to the activations of penultimate layer of a pre-trained Sketch-a-Net \cite{sketchanet2} classifier. Figure.~\ref{fig:lenhist}(d) shows that our model can generate long sketches better than sequential models like SketchRNN in terms of FID (lower is better).

\keypoint{Quantitative Analysis on Sketch Vectorization Quality}
We finally validate our model as a raster-to-curve vectorizer. We vectorize testing sketches and calculate the test loss between vectorized sketch and the available stroke-segmented ground-truth to evaluate the quality of sketch curve fitting. Furthermore, to evaluate the vectorization quality from a semantic perspective, we calculate classification accuracy of the generated vector sketches by first rasterizing and then classifying them with a pre-trained Sketch-a-Net \cite{sketchanet2}. The results are shown in Table~\ref{tab:quant}. We notice that although the usage of sequential information (MSE loss) helps reducing the SWD error a lot, the perceptual quality (classification score) remains fairly similar. \ayandas{We also evaluated B\'ezierSketch \cite{das2020bziersketch} on \quickdraw{}, with and without using stroke sequence information, which correspond to an upper bound and a baseline respectively to our contribution of cloud-based sketch-rendering. The results show that Cloud2Curve almost matches B\'ezierSketch with full sequence supervision, and is significantly better than B\'ezierSketch using raw cloud data (and random sequence assignment).}


\begin{table}
\centering
\scriptsize
\begin{tabular}{r|cc|cc}
\hline
 & \multicolumn{2}{c|}{QuickDraw} & \multicolumn{2}{c}{KMNIST} \\ \hline 
 & \begin{tabular}[c]{@{}c@{}}Classif.\\ acc.\end{tabular} & \begin{tabular}[c]{@{}c@{}}Test\\ Loss\end{tabular} & \begin{tabular}[c]{@{}c@{}}Classif.\\ acc.\end{tabular} & \begin{tabular}[c]{@{}c@{}}Test\\ Loss\end{tabular} \\ \hline
SWD, $\lambda_d=10^{-3}$ & $0.80$ & $0.00123$ & $0.82$ & $8.4\cdot 10^{-3}$ \\ 
SWD, $\lambda_d=10^{-2}$ & $0.69$ & $0.02850$ & $0.73$ & $7.6\cdot 10^{-2}$ \\
SWD + MSE                & $0.84$ & $0.00034$ & $0.85$ & $2.1\cdot 10^{-4}$ \\ \hline
B\'ezierSketch (Seq.) \cite{das2020bziersketch} & 0.86 & 0.00012 & NA & NA \\
B\'ezierSketch (Cloud) \cite{das2020bziersketch}  & 0.41 & 0.2572 & NA & NA \\
Real data    & $0.92$ & NA        & $0.89$ & NA \\ \hline
\end{tabular}
\vspace*{0.1cm}
\caption{Quantitative validation of the vectorization model.}
\label{tab:quant}
\end{table}

\section{Conclusion}

In this paper, we introduced a model capable of generating scaleable vector-graphic sketches using parametric curves -- and crucially it is able to do so by training on point cloud data, thus being widely applicable to general raster image datasets such as K-MNIST. Our framework provides accurate and flexible fitting due to the ability to chose curve complexity independently for each stroke. Once trained, our architecture can also be used to vectorize raster sketches into flexible parametric curve representations. In future work we will generalize our parametric stroke model from overly smooth B\'ezier curves to more general parametric curves such as B-splines or Hermite splines.


{\small
\bibliographystyle{ieee_fullname}
\bibliography{main}
}

\end{document}